\documentclass[10pt, conference, compsocconf]{IEEEtran}

\usepackage{cite}

\usepackage{times}
\usepackage[pdftex]{graphicx}

\usepackage[cmex10]{amsmath}
\usepackage{amssymb}
\usepackage{amsfonts}
\usepackage{array}

\usepackage[caption=false,font=footnotesize]{subfig}

\usepackage[breaklinks=true, bookmarks=false, hidelinks]{hyperref}
\usepackage{multirow}
\usepackage[table, xcdraw, usenames, dvipsnames]{xcolor}

\DeclareMathOperator*{\argmin}{argmin}
\DeclareMathOperator*{\minimise}{min}

\newcommand{\bb}[1]{\boldsymbol{\mathbf{#1}}}

\let\OLDthebibliography\thebibliography
\renewcommand\thebibliography[1]{
  \OLDthebibliography{#1}
  \setlength{\parskip}{0pt}
  \setlength{\itemsep}{0pt plus 0.3ex}
}

\makeatletter
\def\mycopyrightnotice{
  {\footnotesize
  \begin{minipage}{0.3\textwidth}
  978-1-5386-6481-0/18/\$31.00 \textcopyright2018 IEEE \\
  DOI 10.1109/CRV.2018.00029
  \end{minipage}}%
  \begin{minipage}{0.4\textwidth}
  \hfil%
  \hspace{-1pt}\thepage%
  \hfil%
  \end{minipage}
  \begin{minipage}{0.3\textwidth}
  ~
  \end{minipage}
}

\def\ps@headings{%
\def\@oddfoot{\hfil \thepage \hfil}%
\def\@evenfoot{\hfil \thepage \hfil}}

\def\ps@IEEEtitlepagestyle{%
\def\@oddhead{\hfil 2018 15th Conference on Computer and Robot Vision \hfil}%
\def\@oddfoot{\mycopyrightnotice}%
\def\@evenfoot{\hfil \thepage \hfil}}

\makeatother
\begin{document}
%
\title{Visual Object Tracking: The Initialisation Problem}


\author{\IEEEauthorblockN{George De Ath, Richard Everson}
\IEEEauthorblockA{Department of Computer Science\\
University of Exeter\\
Exeter, United Kingdom\\
\{gd259, r.m.everson\}@exeter.ac.uk}
}

\maketitle

\begin{abstract}
Model initialisation is an important component of object tracking. Tracking 
algorithms are generally provided with the first frame of a sequence and a 
bounding box (BB) indicating the location of the object. This BB may 
contain a large number of background pixels in addition to the object and can 
lead to parts-based tracking algorithms initialising their object models in 
background regions of the BB. In this paper, we tackle this as a
missing labels problem, marking pixels sufficiently away from the BB
as belonging to the background and learning the labels of the unknown pixels.
Three techniques, One-Class SVM (OC-SVM), Sampled-Based Background Model (SBBM) 
(a novel background model based on pixel samples), and Learning Based Digital 
Matting (LBDM), are adapted to the problem. These are evaluated with 
leave-one-video-out cross-validation on the VOT2016 tracking benchmark. Our 
evaluation shows both OC-SVMs and SBBM are capable of providing a good level of
segmentation accuracy but are too parameter-dependent to be used in real-world 
scenarios. We show that LBDM achieves significantly increased performance with 
parameters selected by cross validation and we show that it is robust to 
parameter variation.
\end{abstract}


\section{Introduction}
Object tracking in video is an important topic within computer vision, with
a wide range of applications, such as surveillance, activity analysis, robot
vision, and human-computer interfaces. Recent benchmarks for this problem have
been created to evaluate algorithms that are designed for single-object, 
short-term, and model-free, tracking \cite{OTB2013, VOT2016}. The model-free
aspect of the problem is particularly challenging as this means that a tracker
has no prior knowledge of the characteristics (such as colour, shape, and
texture) of the object. In common with real world applications, tracking
algorithms are only provided with the first frame of a video, along with a
bounding box (BB) that indicates which region of the image contains the object.
Typically up to 30\% of this region is comprised of pixels not belonging to the
object, i.e.\ background \cite{VOT2016}. Initialising tracking algorithms with
background instead of foreground can be severely deleterious to their
performance, so in this paper we examine the initialisation problem of locating
the object to be tracked within the given BB.

Tracking algorithms generally fall into two categories: those that track the
entire object via some holistic model which captures the appearance of the
object and potentially its surroundings in some way \cite{KCF, MDNET, STRUCK},
and parts-based models \cite{BHMC, LGT, DPCF}. These latter decompose an object
into a loosely connected set of parts, each with its own visual model, allowing
for better modelling of objects which undergo geometrical deformations and
changes in appearance \cite{LGT}. Incorrect initialisation of parts-based
trackers can lead to multiple parts being initialised to regions, within the
BB, that do not belong to the object. This can cause parts to drift
away from the object during tracking, as the object, but not its background,
moves.

This problem is typically dealt with in two \textit{ad hoc} ways: attempting to
select regions of the BB to track that are highly likely to belong to
the object; and also removing parts from an object's model that exhibit signs
of poor performance, reinitialising these parts as required \cite{BHMC}. Both of
these methods attempt to ascertain the object's exact location based on the
limited information available to them, knowing only where the object 
approximately resides, such as inside the BB during initialisation.
However, it is not known which pixels within, and close to, this region belong
to the object and which belong to its background. We refer to the problem of
determining which pixels, given an approximate location of the object, belong to
the object and which do not, as the \textit{Initialisation Problem}.

Recently, some parts-based techniques have attempted to address the 
initialisation problem in various ways. Several methods \cite{LGT, DPCF,
elastic_patches} distribute their parts uniformly over the BB and
employ aggressive update schemes to identify patch drift. Other methods, e.g.\
\cite{GGTv2, CDTT}, employ oversegmentation techniques, such as
superpixeling, to mark all pixels belonging to a superpixel that crosses the
BB as not belonging to the object. Others attempt to select regions of
the BB with desirable attributes for the particular tracking algorithm
being used. These include selecting areas likely to have good optical
flow estimation \cite{RVT_TA} and areas with good image alignment properties
\cite{BHMC}.

In this paper we address the initialisation problem by treating it as a missing
labels problem, as we can be sure that pixels sufficiently outside the BB
do not belong to the object. The problem is challenging, however, because
BBs are generally small and the object itself may be quite similar in
appearance to the background. In addition, the object to be tracked may extend
outside the given BB \cite{VOT2016_seg}. For example, in order to
reduce the number of background pixels within a BB, the outstretched
limb of a person to be tracked might be excluded from the initialising BB. 
This means that the distance away from a BB where pixels are
certainly background may be different for each segmentation.  We note too that
distance from the BB is not a guarantee that a pixel or region's
appearance is not similar to the appearance of the object to be tracked; a
common example is tracking a particular face in a crowd.

We evaluate the performance of three techniques: the first two, One-Class SVM
(OC-SVM) and a novel Sample-Based Background Model (SBBM), treat the problem as
a one-class classification problem. They attempt to learn the characteristics of
the background regions of the image; pixels within the BB are then
compared to the background model to determine object pixels. The third technique
is an adapted solution to the image matting problem. This models each pixel in
the image as comprising proportions of both foreground and background colour,
aiming to to learn these proportions. The performance of these three techniques is
evaluated by assessing their segmentation on the Visual Object Tracking 2016,
VOT2016, dataset \cite{VOT2016}. We discuss their strengths and weaknesses and
investigate the robustness to parameter settings of the learned matting method.

The rest of the paper is organised as follows. In Section \ref{sec:methods} we
formulate the initialisation problem and present the three techniques to be
empirically compared. The experimental procedure and results are presented in
Sections \ref{sec:evaluation-protocol} and \ref{sec:results}, and are followed
by concluding remarks in Section \ref{sec:conclusion}.

\section{Methods}
\label{sec:methods}
We aim to determine the label $k \in \left\{0, 1\right\}$ of each pixel in the 
image, where the two labels represent belonging to the background of the 
BB and the object itself, respectively. The image $I$ containing the
object consists of a set of pixels indexed by $\Omega = \left\{ 1, \ldots, n 
\right\}$, where $n$ is the total number of pixels in the image. Given a subset
of these pixels $\Omega_b \subset \Omega$ which belong to the background, such
that $k_i = 0 \: \forall \: i \in \Omega_b$, then the problem can be formulated
as learning the correct labels of the unlabelled pixels $\Omega_u = \Omega
\setminus \Omega_b$.

Rather than treating each pixel $i \in \Omega$, or features extracted from them,
as a data point, two of the methods we examine use a superpixel representation,
which provides a richer description of small, approximately homogeneous, image
regions. Superpixeling techniques oversegment the image into perceptually
meaningful regions (the pixels in a superpixel are generally uniform in colour
and texture), while at the same time retaining the image structure, as 
superpixels tend to adhere to colour and shape boundaries. Using superpixeling 
also can also significantly reduce the computational complexity because groups of
pixels are represented as a single entity. An image $I$ is therefore comprised
of a set of superpixels $\mathcal{S} = \left\{ S_j \right\}_{j=1}^{{N_{sp}}}$,
where $N_{sp}$ is the number of superpixels, and we associate a vector of
features $\bb{x}_j \in \mathbb{R}^d$, such as a histogram of its RGB values, 
with each superpixel.

\subsection{One-Class SVM}

In general, pixels belonging to the object are not known, but a large number
of pixels or superpixels may be assumed to belong to the background. We 
therefore treat the initialisation problem as that of identifying pixels which
are significantly different from the background. This can be accomplished by
using a one-class SVM (OC-SVM) to estimate the support of the given background
pixels.

The goal of an OC-SVM \cite{OCSVM_Scholkopf_TR} is to learn the hyperplane 
$\bb{w} \in \mathcal{F}$ that produces the maximum separation between the origin
and data points $\{ \bb{x}_i \}_{i=1}^N, \, \bb{x}_i \in \mathcal{X}$ mapped to
a suitable feature space $\mathcal{F}$. It uses an implicit function 
$\Phi\left(\cdot\right)$ to map the data into a dot product feature space 
$\Phi$: $\mathcal{X} \mapsto \mathcal{F}$ using a kernel 
$k\left(\bb{x},\bb{y}\right) = \langle \Phi\left(\bb{x}\right),
\Phi\left(\bb{y}\right) \rangle$. The radial basis function kernel  (Gaussian
kernel), $k\left(\bb{x},\bb{y}\right) = e ^{ - \gamma \left\lVert \bb{x} -
\bb{y} \right\rVert^2}$, is a popular choice, as this guarantees the existence
of such a hyperplane \cite{OCSVM_Scholkopf_TR}. The decision function 
$f\left(\bb{x}\right) = \text{sgn}\left(\bb{w} \cdot \Phi\left(\bb{x}\right) -
\rho \right)$, is used to determine whether an arbitrary input vector $\bb{x}$
belongs to the one defined class when $\bb{w} \cdot \Phi\left(\bb{x}\right)
\ge \rho$.

In order to find $\bb{w}$ and the threshold $\rho$, the problem can be defined 
in its primal form:
\begin{IEEEeqnarray}{rc}
\minimise_{\bb{w} \in \mathcal{F}, \: \bb{\xi} \in \mathbb{R}^N, \: \rho \in \mathbb{R}} & \frac{1}{2} \left\lVert \bb{w} \right\rVert^2 - \rho + \frac{1}{\nu N} \sum_i \xi_i \label{eqn:ocsvm_primal_obj} \\
\text{subject to}
& \quad \bb{w} \cdot \Phi\left(\bb{x}_i\right)  \ge \rho - \xi_i, \quad \xi_i \ge 0,
\end{IEEEeqnarray}
where slack variables $\xi_i$ allow the corresponding $\bb{x}_i$ to lie on the 
other side of the decision boundary, and $\nu \in \left(0, 1 \right)$ is a 
regularisation parameter. Conversion to the dual form reveals that the solution
can be written as a sparse weighted sum of \textit{support vectors} and solved
by quadratic programming or, more typically, Sequential Minimal Optimisation
\cite{LIBSVM}.

The parameter $\nu$ acts both as an upper bound on the fraction of outliers
(data points with $f\left(\bb{x}\right) < 0$) and a lower bound on the fraction
of data points used as support vectors. It is worth noting that, for kernels
such that $k(\bb{x}_i, \bb{x}_i) = c \: \forall \: \bb{x}_i \in \mathcal{X}$,
where $c$ is some constant, this formulation is equivalent \cite{LwK} to that of
using Support Vectors for Data Description \cite{OCSVM_SVDD}, in which the goal
is to encapsulate most of the data in a hypersphere in $\mathcal{F}$.

\subsubsection{Application}
\label{sec:ocsvm-application}

\begin{figure}[t]
\centering
\subfloat[]{\fbox{%
\includegraphics[width=0.116\textwidth, clip, trim={0 10 0 10}]{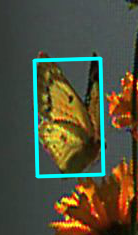}%
\label{fig:ocsvm_example_1}}}
\hfil
\subfloat[]{\fbox{%
\includegraphics[width=0.116\textwidth, clip, trim={0 10 0 10}]{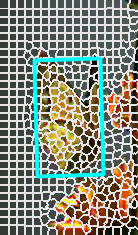}%
\label{fig:ocsvm_example_2}}}
\hfil
\subfloat[]{\fbox{%
\includegraphics[width=0.116\textwidth, clip, trim={0 10 0 10}]{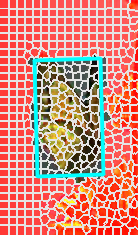}%
\label{fig:ocsvm_example_3}}}
\hfil
\subfloat[]{\fbox{%
\includegraphics[width=0.116\textwidth, clip, trim={0 10 0 10}]{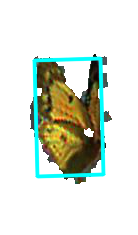}%
\label{fig:ocsvm_example_4}}}
\caption{The OC-SVM pipeline. The image is first cropped (a) and then 
superpixeled (b). The OC-SVM is trained ($\gamma = 2^{-20}$, $\nu = 0.3$) using
the superpixels considered  \textcolor{red}{background} in (c), and those with
an unknown label (uncoloured) have their class predicted, giving a final 
predicted object mask (d).}
\label{fig:ocsvm_example}
\end{figure}

In order to use the OC-SVM approach, the image is preprocessed and features 
extracted in order to train the classifier and identify which regions of the 
image belong to the object. As illustrated in Fig.~\ref{fig:ocsvm_example_1},
the image $I$ is first cropped to twice the width and height of an axis-aligned
BB containing the supplied BB and object. The cropping allows
for a sufficient region of the image to be included to train the classifier,
while only including areas relatively close, in terms of the object's size, to
the object. The cropped region is then segmented using a variant of the SLIC
superpixeling algorithm \cite{SLIC}, SLIC0, which produces regular shaped 
(compact) superpixels (Fig.~\ref{fig:ocsvm_example_2}), while adhering to image
boundaries in both textured and non-textured regions of the image. 

The main parameter of SLIC0 is the approximate number of superpixels $N_{sp}$
resulting from the segmentation. Following preliminary experiments, we aim to
have an average of 50 pixels per superpixel to give a sufficient number of
pixels from which to extract features. However, given the varying size of each
cropped image, this could result in very few or very many superpixels. Therefore
we constrain the number of superpixels to lie in the range $\left[N_{sp}^-,
N_{sp}^+\right]$. We use
$N_{sp}^- = 100$ and $N_{sp}^+ = 500$, as these have empirically given reasonable-looking
segmentations and are similar to the range used by \cite{GGTv2}.

Following segmentation, superpixels that lie wholly outside the BB are 
labelled as not belonging to the object, such as the red superpixels in 
Fig.~\ref{fig:ocsvm_example_3}, with the features extracted from these forming
the training data. We examine the performance of four alternative feature
representations: histograms of RGB or, perceptually uniform, LAB pixel
intensities across the superpixel; or the concatenated SIFT \cite{SIFT} or LBP
\cite{LBP} feature vectors resulting from the R, G and B channels at the
centroid of the superpixels. Both SIFT and LBP are popular texture-based feature
descriptors, with SIFT features extracted based on the region's gradient
magnitude and orientation \cite{SIFT}, and LBP features formed based on
differences in intensity between a pixel and its surrounding neighbours
\cite{LBP}.

The classifier, with its parameters  $\gamma$ and $\nu$, representing the
kernel's length-scale and the classifier's upper bound on the assumed number of
outliers in the training set, is subsequently trained on this data. Finally,
features extracted from the unknown superpixels, lying wholly or partially
within the given BB are classified with the trained OC-SVM. 
Fig.~\ref{fig:ocsvm_example_4} shows the resulting segmentation.

\subsection{Sample-Based Background Model}
\begin{figure}[t]
\centering
\subfloat[]{\fbox{
\includegraphics[width=0.116\textwidth, clip, trim={0 70 0 70}]{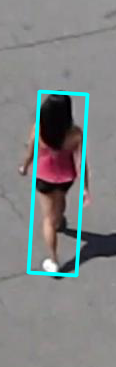}%
\label{fig:vibelike_example_1}}}
\hfil
\subfloat[]{\fbox{%
\includegraphics[width=0.116\textwidth, clip, trim={0 70 0 70}]{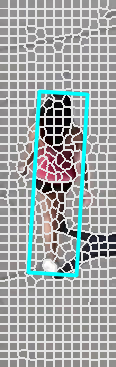}%
\label{fig:vibelike_example_2}}}
\hfil
\subfloat[]{\fbox{%
\includegraphics[width=0.116\textwidth, clip, trim={0 70 0 70}]{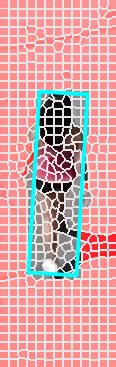}%
\label{fig:vibelike_example_3}}}
\hfil
\subfloat[]{\fbox{%
\includegraphics[width=0.116\textwidth, clip, trim={0 70 0 70}]{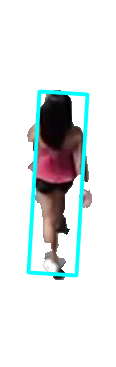}%
\label{fig:vibelike_example_4}}}
\caption{The SBBM pipeline. The image is cropped (a) and superpixeled (b). 
A model ($\delta = 0.5$) is created for each \textcolor{red}{superpixel}
in (c) outside the BB, and the other superpixels (uncoloured)
are compared to them ($\eta = 0.8$) and classified (d).}
\label{fig:vibelike_example}
\end{figure}

Our sample-based one-class modelling technique is inspired by the background
subtraction algorithm of \cite{VIBE}. We represent labelled regions of the
image, in this case superpixels, by sets of samples that characterise the colour
distribution of these regions. In the same way as described in Section
\ref{sec:ocsvm-application}, the region surrounding the supplied BB is
cropped and over-segmented into superpixels (Fig.~\ref{fig:vibelike_example_1}
and \ref{fig:vibelike_example_2}). Superpixels lying wholly outside the
specified BB are labelled as background 
(Fig.~\ref{fig:vibelike_example_3}).

An unlabelled superpixel is compared to the modelled superpixels by evaluating
how similar its set of samples is to each of the labelled sets of samples. The
unlabelled superpixel is classified as either belonging to the labelled region
(background) of the image if it is sufficiently similar to any of the labelled
sets of samples, or the object if not. We denote the model of the $j$-th
superpixel $S_j$ to be $m_j = \{ \bb{x}_i \}_{i=1}^s$, consisting of a set of
$s$ pixel values $\bb{x}_i$ randomly sampled (with replacement) from $S_j$. This
can be thought of as an empirical histogram of the superpixel's colour
distribution. As the average number of pixels in a superpixel, $\bar{N}_p$, may
vary from image to image, we set $s = \delta \bar{N}_p$, where $\delta \in (0,
1]$ is chosen by cross-validation.

Let $\mathcal{M} = \{ m_j \}$ be the set of models that characterise the 
superpixels which are located completely outside the BB. Then pixel 
$\bb{x}_p$ in a superpixel whose model is $m' = \{ \bb{x}_p \}_{p=1}^s$, is
deemed to match the model $m_j$ if it is closer than a radius $R$ to any pixel in 
$m_j$. Thus
\begin{equation}
Q\left( \bb{x}_p, m_j \right) = 
\begin{cases}
1 & \exists \: 
         \bb{x}_i \in m_j: \left\lVert \bb{x}_i - \bb{x}_p \right\rVert_2 < R \\
0 & \text{otherwise.}
\end{cases}
\end{equation}
The parameter $R$ controls the radius of a sphere centred on each of the model's
pixels, allowing for inexact matches. This is needed as lighting conditions
usually vary within an image, so it permits matching pixels with subtle 
discrepancies in colour which would otherwise result in a mismatch.

The extent to which pixels in $m'$ match $m_j$ is then assessed by
\begin{equation}
q(m', m_j) = \frac{1}{S} \sum_{\bb{x}_p \in m'}^S Q\left(\bb{x}_p, m_j\right).
\end{equation}
If $q(m', m_j) > \eta$ for any $m_j \in \mathcal{M}$ then the unlabelled
superpixel is marked as matching a background superpixel and is deemed to be
itself a background superpixel. If $m'$ fails to match any background superpixel
it is counted as foreground. Comparison of $q(m', m_j)$ with $\eta $ allows for
some false positive matches before the superpixel is counted as background. 

Fig.~\ref{fig:vibelike_example} illustrates the pipeline. We remark that SBBM 
has correctly identified the majority of the shadows within the BB as 
background, and the person's shoes as foreground. However, parts of their arms
and legs have been labelled as background because they match limb or shadow 
pixels outside the BB.

\subsection{Learning Based Digital Matting}
Digital matting, also known as natural image matting \cite{LBDM}, is the process
of separating an image $I$ into a foreground $F$ and background $B$ image, along
with an opacity mask $\alpha$. The colour of the $i$-th pixel is assumed to be a
combination of a corresponding foreground and background colour, blended
linearly,
\begin{equation}
I_i = \alpha_i F_i + \left( 1 - \alpha_i \right) B_i,
\end{equation}
where $\alpha_i \in [0, 1]$ is the pixel's foreground opacity. Solving for 
$\alpha_i$, $F_i$, and $B_i$ is extremely under-constrained as there are more 
unknown $\left( F, B, \alpha \right)$ than known $\left( I \right)$ variables. 
This ill-posed problem is made more tractable by supplying additional 
information, using either trimaps, where the vast majority of the image is 
labelled as being all foreground or all background pixels (e.g.\ 
\cite{BAYESIAN_MATTING}), or by using scribbles, where only small regions, 
denoted by user-supplied scribbles, are selected as being foreground or 
background.

Learning Based Digital Matting \cite{LBDM} trains a local alpha-colour model for 
all pixels $\Omega$ based on their most-similar neighbouring pixels. More 
specifically, any pixel's $i \in \Omega$ corresponding alpha matte value 
$\alpha_i$ is predicted via a linear combination of the alpha values 
$\{\alpha_j \}_{j \in \mathcal{N}_i}$, where $\mathcal{N}_i \subset \Omega$ are
the neighbouring pixels of $i$. Defining  $\bb{\alpha}_i = [ \alpha_{\tau_1},
\ldots, \alpha_{\tau_j}, \ldots, \alpha_{\tau_m} ]^T$, with $\tau_j \in 
\mathcal{N}_i$ and $m = |\mathcal{N}_i|$, to be the $\alpha$ values of the
neighbouring pixels, the linear combination is expressed as 
\begin{equation}
\alpha_i = \bb{f}_i^T \bb{\alpha}_i, \label{eqn:alpha_matting_pixel}
\end{equation}
where $\bb{f}_i = [f_{i\tau_1}, \ldots, f_{i\tau_j}, \ldots, f_{i\tau_m}]^T$
are the coefficients of the alpha values.

Equation \eqref{eqn:alpha_matting_pixel} can also be written in terms of a linear 
combination of the alpha values for the entire image: $\bb{\alpha} = [\alpha_1, 
\ldots, \alpha_n]^T$. Defining the $n$ by $n$ matrix $\bb{F}$ such that the 
$i$-th column contains the coefficients $\bb{f}_i$ in positions corresponding 
to $\mathcal{N}_i$, $\bb{\alpha}$ can be estimated via a mean squared error
minimisation
\begin{equation}
\bb{\alpha}^* = \argmin_{\bb{\alpha}} \quad  \lVert \bb{\alpha} - \bb{F}^T
\bb{\alpha} \rVert^2 + c \lVert \bb{\alpha}_b - \bb{\alpha}_b^{*} \rVert,
\label{eqn:alpha_matting_minimisation}
\end{equation}
where $\bb{\alpha}_b \subset \bb{\alpha}$ and $\bb{\alpha}_b^{*}$ are vectors of the predicted alpha 
values of the labelled pixels and the actual alpha values of labelled pixels
respectively. The parameter $c$ denotes the size of the penalty applied for
predicting alpha values that are different than the user-specified labels;
\cite{LBDM} set $c = \infty$ in order to penalise any deviation and to maximally
use the additional information provided by the known labels.

Equation \eqref{eqn:alpha_matting_minimisation} can also be written as
\begin{equation}
\setlength{\jot}{-4pt} 
\begin{split}
\bb{\alpha}^* = \argmin_{\bb{\alpha} \in \mathbb{R}^n} \quad & 
\bb{\alpha}^T \left( \bb{I} - \bb{F} \right) \left( \bb{I} - \bb{F} \right)^T
\bb{\alpha} \\
& + \left( \bb{\alpha} - \hat{\bb{\alpha}} \right) ^ T \bb{C} \left( \bb{\alpha} - \hat{\bb{\alpha}} \right),
\end{split}
\end{equation}
where $\bb{I}$ is the identity matrix, $\bb{C}$ is a diagonal matrix with 
$C_{ii} = c$ if $i \in \Omega_b$ and 0 otherwise, and  $\hat{\bb{\alpha}}$ is
the vector whose elements are the provided alpha values for  $i \in \Omega_b$
and zero otherwise. Assuming $\bb{F}$ is known, this is solved by
\begin{equation}
\bb{\alpha}^* = \left( \left( \bb{I} - \bb{F} \right) \left( \bb{I} - \bb{F} \right)^T + \bb{C} \right)^{-1} \bb{C} \hat{\bb{\alpha}}.
\label{eqn:alpha_matting_thingtosolve}
\end{equation}

The columns of $\bb{F}$ are computed via a local learning model to predict the
value of $\alpha_i$. A linear local colour model for each pixel $i \in \Omega$
is trained to describe the relationship between the alpha values of a data point
and its neighbours. Solving a ridge regression problem \cite{LBDM}, shows that
the non-zero values of the $i$-th column of $\bb{F}$ are only dependent on the
features of each pixel and can be expressed as
\begin{equation}
\bb{f}_i = \left( \bb{X}_i \bb{X}_i^T + \lambda \bb{I} \right)^{-1} \bb{X}_i \bb{x}'_i,
\end{equation}
where $\bb{X}_i = [\bb{x}_{\tau_1} \ldots \bb{x}_{\tau_m}, \bb{1} ]^T$ is a
matrix populated by the data points in the neighbourhood of pixel $i$, and 
$\bb{x}'_i = [ \bb{x}^T, 1 ]^T$. The shrinkage parameter $\lambda$ controls
the size of the penalty placed on the regularisation coefficients.

\subsubsection{Application}
In order to apply this algorithm to the initialisation problem a 
\textit{scribble mask} is needed. This contains labels for pixels that 
definitely belong to the background, the object, and the unknown region. Alpha 
matting typically has a scribble mask provided  via user input, however this is
not possible within the object tracking paradigm, as no additional \textit{a
priori} information can be provided.

\begin{figure}[t]
\centering
\subfloat[]{\fbox{
\includegraphics[width=0.155\textwidth, clip, trim={0 15 0 20}]{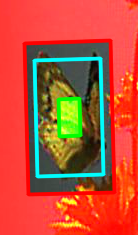}%
\label{fig:alpha_matting_example_1}}}
\hfil
\subfloat[]{\fbox{%
\includegraphics[width=0.155\textwidth, clip, trim={0 15 0 20}]{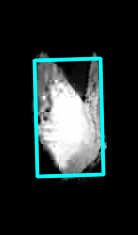}%
\label{fig:alpha_matting_example_2}}}
\hfil
\subfloat[]{\fbox{%
\includegraphics[width=0.155\textwidth, clip, trim={0 15 0 20}]{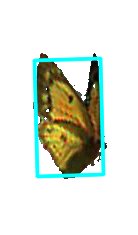}%
\label{fig:alpha_matting_example_3}}}
\caption{The alpha matting pipeline. (a) The \textcolor{cyan}{original}, 
\textcolor{red}{expanded} ($\rho^+ = 1.7$), and \textcolor{ForestGreen}{contracted} ($\rho^- = 
0.9$) BBs. The calculated alpha matte (b) and the mask  (c) 
corresponding to a threshold of $\tau = 0.80$.}
\label{fig:alpha_matting_example}
\end{figure}

We have addressed this by automating the creation of a scribble mask based on 
only the BB and image, cropped in an identical manner to that of the 
superpixeling algorithms. The area of the original BB (cyan box in 
\ref{fig:alpha_matting_example_1}) is decreased by a factor of $\rho^- \in (0, 
1)$, linearly shrinking the height and width of the BB by 
$\sqrt{\rho^-}$. The pixels within this region are labelled in the scribble mask 
as belonging to the object, shown by the green-shaded area in 
Fig.~\ref{fig:alpha_matting_example_1}. Similarly, we also increase the area of
the original BB by a factor of $\rho^+ \in (1, 2]$, linearly expanding 
both of its dimensions by $\sqrt{\rho^+}$. Pixels outside this region are 
labelled as belonging to the background (shaded red in 
Fig.~\ref{fig:alpha_matting_example_1}), leaving pixels between the two labelled 
regions as being of unknown origin.

Shrinking the area of the BB and using this region as a scribble mask
to represent the object relies on the assumption that the object is located at
the centre of the BB, which may not always be the case. If an object
is highly non-compact or has holes in it, then there is a possibility of 
background pixels being included in the scribble mask. Expanding the BB 
to represent the background also makes the assumption that pixels 
sufficiently far away from the BB do not belong to the object. This is 
a stronger assumption, although there may still be cases where parts of
an object protruding far from the BB are labelled as background.

The output of the alpha matting process, the alpha matte $\bb{\alpha}^*$, is not 
a strict segmentation, but rather a pixel mask indicating what fraction of each
pixel is foreground (belonging to the object) and background. A segmentation is
produced by thresholding the alpha values to create an object mask. Pixels with
a predicted $\alpha_i^*$ value greater than a threshold $t$ are assigned as
belonging to the object, with those whose $\alpha_i \leq t$ being labelled as
background. The threshold $t$ is chosen so that a proportion $\tau$ of the
BB is classified as the object. This allows the alpha threshold $t$ to
be dynamically chosen based on how much of the BB is expected to be
populated with the object in question. For example, in the VOT2016 competition
\cite{VOT2016} at least 70\% of the BB belongs to the object.

\section{Evaluation Protocol}
\label{sec:evaluation-protocol}
In order to evaluate the segmentation performance of each algorithm, we used
real-world input in the form of the VOT2016 dataset \cite{VOT2016}. It is one of
the most widely used datasets in the tracking community, comprising  60 
videos, with human-annotated ground-truth BBs and pixel-wise 
segmentations \cite{VOT2016_seg}. In order to increase the number of examples,
we used the first, middle, and last frames from each video. Within this dataset,
the sizes of the BBs varies widely. The minimum and maximum 
dimensions were $8.5$ and $511$ pixels respectively, with the mean dimension
being $92$ pixels. The number of pixels contained within the BBs
ranged from $229$ to $214,773$ with the mean being $12,183$.

The segmentation performance was assessed by the Intersection over Union (IoU) 
measure, also known as the Jaccard index:
\begin{equation}
  \label{eq:IoU}
\phi_{all} = \frac{ \left| \mathcal{G} \cap \mathcal{P} \right| }
                  { \left| \mathcal{G} \cup \mathcal{P} \right| },
\end{equation}
where $\mathcal{G}$ and $\mathcal{P}$ are the sets of pixels forming the 
ground-truth and predicted segmentations. It is worth noting that the IoU 
measure is equivalent to the Dice similarity coefficient (DSC), also known as 
the $\text{F}_1$ measure, in the sense that 
$\text{DSC} = 2 \phi_{all} / ( 1 + \phi_{all} )$.

\begin{figure}[t]
\centering
\subfloat[]{\fbox{
\includegraphics[width=0.155\textwidth, clip, trim={0 30 0 0}]{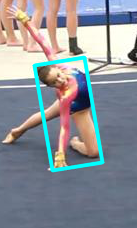}%
\label{fig:seg_assessment_crit_1}}}
\hfil
\subfloat[]{\fbox{%
\includegraphics[width=0.155\textwidth, clip, trim={0 30 0 0}]{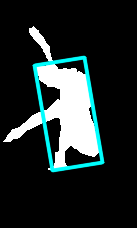}%
\label{fig:seg_assessment_crit_2}}}
\hfil
\subfloat[]{\fbox{%
\includegraphics[width=0.155\textwidth, clip, trim={0 30 0 0}]{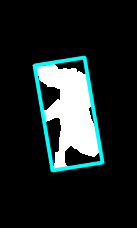}%
\label{fig:seg_assessment_crit_3}}}
\caption{(a) The original cropped image and BB. Ground truths masks of 
the (b) full $\phi_{all}$ and (c) BB $\phi_{bb}$ segmentation 
criterion.}
\label{fig:seg_assessment_crit}
\end{figure}

The segmentations were assessed using two criteria, $\phi_{all}$ and 
$\phi_{bb}$. As illustrated in Fig.~\ref{fig:seg_assessment_crit_2}, the 
$\phi_{all}$ measure compares the predicted object mask $\mathcal{P}$ to the 
ground truth $\mathcal{G}$, including regions of $\mathcal{G}$ that lie
outside the BB. This assesses the full segmentation capability of the
technique, and allows a tracker to make use of all available information about
the object and its background. The $\phi_{bb}$ measure compares the quality of
object segmentation and the ground truth within the BB, $\mathcal{B}$;
thus 
\begin{equation}
  \label{eq:IoU_bb}
\phi_{bb} = \frac{ \left| \mathcal{G} \cap \mathcal{P} \cap \mathcal{B} \right| }
            { \left| (\mathcal{G} \cup \mathcal{P}) \cap \mathcal{B}
              \right| }.
\end{equation}
This corresponds to typical use by tracking algorithms which assume that the
object lies completely within the BB, and only track the pixels within
it.

\begin{table}[t]
\renewcommand{\arraystretch}{1.15}
\centering
\caption{Parameter values used in cross-validation.}
\label{tbl:params}
\begin{tabular}{|c|c|l|}
\hline
\textbf{Technique} & \multicolumn{1}{c|}{\textbf{Parameter}} & \multicolumn{1}{c|}{\textbf{Values}} \\ \hline
\multirow{3}{*}{OC-SVM} & $\nu$     & \begin{tabular}[c]{@{}l@{}}
							 	       $0.001, 0.002, \ldots, 0.005,$ \\
									   $0.01, 0.02, \ldots, 0.05,$ \\ 
                                       $0.1, 0.15, \ldots, 0.5$
                                      \end{tabular} \\ \cline{2-3}
                        & $\gamma$  & $2^{-20}, 2^{-19}, \ldots, 2^{20}$ \\ \hline
\multirow{2}{*}{SBBM}   & $\delta$  & $0.1, 0.2, \ldots, 1.0$ \\ \cline{2-3} 
                        & $\eta$    & $0.1, 0.2, \ldots, 1.0$ \\ \hline
\multirow{4}{*}{LBDM}   & $\rho^+$  & $1.1, 1.2, \ldots, 2.0$ \\ \cline{2-3} 
                        & $\rho^-$  & $0.1, 0.2, \ldots, 0.9$ \\ \cline{2-3} 
                        & $\tau$    & $0.50, 0.51,  \ldots, 1.00$ \\ \cline{2-3} 
                        & $\lambda$ & $10^0, 10^{-1}, \ldots, 10^{-10}$ \\ \hline
\end{tabular}
\end{table}

Each technique was evaluated using leave-one-out cross-validation based on
videos, meaning that the 3 frames from a single video were held out for testing,
while the optimal parameters for the technique were determined as those giving
the best $\phi_{all}$ or $\phi_{bb}$ score averaged over the 177 frames from the
remaining 59 videos. Performance on the 3 held out frames was then evaluated
using the cross-validated optimal parameters. This procedure was repeated for
each of the other 59 videos held out in turn to obtain average performance
metrics.

Each possible combination of the parameter values shown in Table 
\ref{tbl:params} was evaluated for each technique, with the process being
repeated for both $\phi_{all}$ and $\phi_{bb}$. In addition, a baseline 
performance, denoted as ``Entire BB", was obtained as the segmentation that
consists of the entire BB; that is $\mathcal{P} \equiv \mathcal{B}$.

As noted above, the OC-SVM is a feature-based classifier. RGB, LAB, SIFT 
\cite{SIFT}, and LBP \cite{LBP} features, with the first two being colour-based 
and the latter two texture-based features, were used for separate experiments 
within the same testing protocol. RGB histograms are one of many representations
of colour data, describing the underlying colour distribution of the image
region as seen by physical devices. Unlike RGB, LAB is designed to approximate
human vision and is perceptually uniform, i.e.\ changes in colour values result in
the same amount of perceived visual change. We used 8 bins for each channel in
both the RGB and LAB feature descriptors to represent each superpixel. SIFT
features were the concatenation of SIFT features extracted for each colour
channel separately; likewise for LBP. Dense SIFT features were extracted with a
$4 \times 4$ sliding window and with the orientation set to 0. The feature
descriptor extracted nearest to a superpixel's centroid was assigned as its
feature. LBP features from a $5 \times 5$ window were extracted, centred on the
pixel nearest to the superpixel's centroid. We used the typical parameters of
$P=8$ and $R=2$, with the scale and rotation invariant version of the
descriptor, along with the uniform patterns extension \cite{LBP}. Features
were standardised to have zero mean and unit variance before classification.

SBBM uses the raw RGB values of the pixels and a colour radius $R=20$,
corresponding to approximately a $4.5\%$ deviation in each colour channel, which
we have found to give good results across a range of examples.

\begin{table}[t]
\renewcommand{\arraystretch}{1.15}
\centering
\caption{Average segmentation performance of each technique for the two overlap
measures. The first and second place values are highlighted by red and blue
colour respectively.}
\label{tbl:results}
\begin{tabular}{l*{2}{|>{\centering\arraybackslash}p{6em}}|}
\cline{2-3}
                                         & \multicolumn{2}{c|}{\textbf{Average Performance}}        \\ \hline
\multicolumn{1}{|l|}{\textbf{Technique}} & $\phi_{all}$            & $\phi_{bb}$              \\ \hline
\multicolumn{1}{|l|}{Entire BB}          & 0.579                   & 0.747                   \\ \hline
\multicolumn{1}{|l|}{OC-SVM + RGB}       & \textcolor{blue}{0.588} & 0.744                   \\ \hline
\multicolumn{1}{|l|}{OC-SVM + LAB}       & 0.581                   & 0.745                   \\ \hline
\multicolumn{1}{|l|}{OC-SVM + SIFT}      & 0.579                   & 0.747                   \\ \hline
\multicolumn{1}{|l|}{OC-SVM + LBP}       & 0.580                   & \textcolor{blue}{0.748} \\ \hline
\multicolumn{1}{|l|}{SBBM}               & 0.555                   & 0.747                   \\ \hline
\multicolumn{1}{|l|}{Alpha Matting}      & \textcolor{red}{0.763}  & \textcolor{red}{0.804}  \\ \hline
\end{tabular}
\end{table}

In the alpha matting technique, the neighbourhood $\mathcal{N}_i$ was defined to
be a $3 \times 3$ region centred on the pixel in question, meaning that the size
of each pixel's neighbourhood was $m=8$. We used the RGB values to be the
features $\bb{x}_i$ of each pixel, normalising them such that $[0, 255] \mapsto
[0, 1]$. The alpha matting parameter was set to $c=800$ as this is a large
enough penalty, compared with normalised values of $\bb{x}_i$, to act
effectively as $\infty$ \cite{LBDM}.

\section{Results}
\label{sec:results}

\begin{figure}[t]
\centering
\subfloat[]{
\includegraphics[width=0.47\textwidth, clip, trim=8 10 8 10]{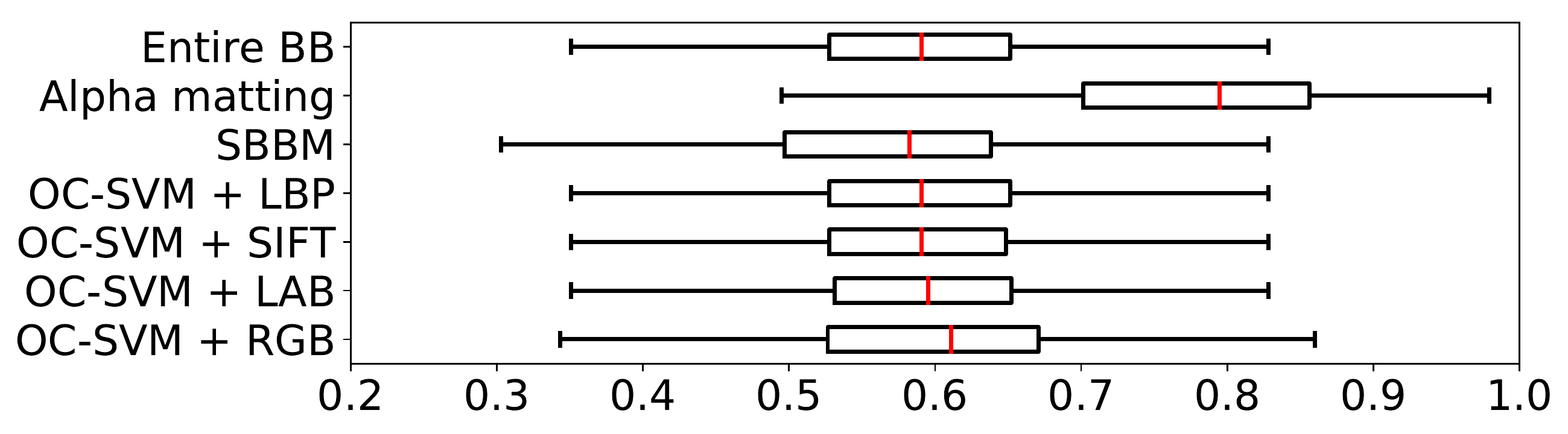}%
\label{fig:performance_box_whisker_all}}%
\\[-1pt]
\subfloat[]{%
\includegraphics[width=0.47\textwidth, clip, trim=8 10 8 10]{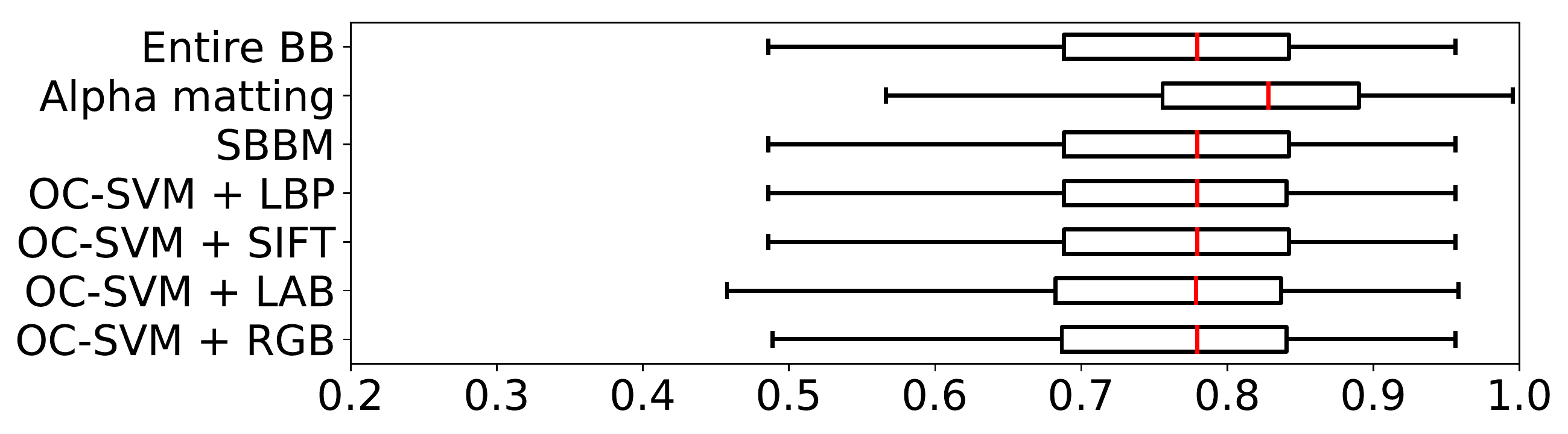}%
\label{fig:performance_box_whisker_bb}}
\caption{Distribution of (a) $\phi_{all}$ and (b) $\phi_{bb}$ 
  for each technique.}
\label{fig:performance_box_whisker}
\end{figure}

Table \ref{tbl:results} gives summary segmentation performance results for both 
$\phi_{all}$ and $\phi_{bb}$, with Fig.~\ref{fig:performance_box_whisker} 
displaying the distribution of scores across the VOT2016 data. The performance
of simply predicting the entire BB to be the object, labelled as
Entire BB in the figures, acts as a baseline and we note that on average the 
OC-SVM and SBBM methods perform similarly to the baseline. The range of scores
is large for all methods (Fig.~\ref{fig:performance_box_whisker}), indicating
that although each method performs well on some images, there others on which it
performs poorly.

\begin{figure}[t]
\centering
\subfloat[]{\fbox{
\includegraphics[width=0.155\textwidth, clip,
				 trim={0 0 0 11}]{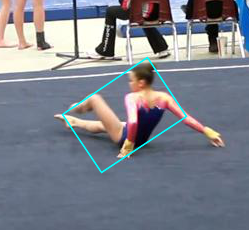}%
\label{fig:results_goodbad_svm_1}}}
\hfil
\subfloat[]{\fbox{%
\includegraphics[width=0.155\textwidth, clip,
				 trim={45 45 45 45}]{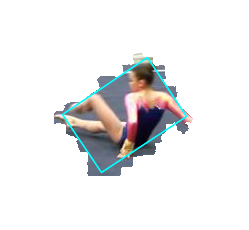}%
\label{fig:results_goodbad_svm_2}}}
\hfil
\subfloat[]{\fbox{%
\includegraphics[width=0.155\textwidth, clip,
				 trim={45 45 45 45}]{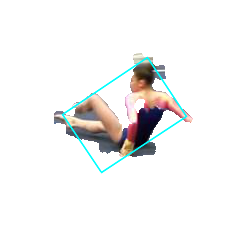}%
\label{fig:results_goodbad_svm_3}}}
\caption{OC-SVM: (a) Original image. (b) Cross validated segmentation
($\gamma = 10^{6}$, $\nu = 0.004$). (c) Best possible segmentation using 
LAB features ($\gamma = 10^{-19}$, $\nu = 0.250$).}
\label{fig:results_goodbad_svm}
\end{figure}

\paragraph{OC-SVM} We compared segmentations using the cross-validated optimal
parameters, with those using the parameters that gave the best result for 
$\phi_{all}$. A typical example of this can be seen in 
Fig.~\ref{fig:results_goodbad_svm}. In general, we found that the OC-SVM was
capable of giving good segmentations, as can be seen in 
Fig.~\ref{fig:results_goodbad_svm_3}, but the parameters required to achieve
these were widely spread, with no one region of parameter space giving good
results for the majority of images.

It is interesting to note that there is little difference between the 
performance of the OC-SVM between the colour and texture-based features. In
addition to the four feature-descriptors reported, we also combined them by 
concatenating the feature vectors together. No statistical improvement was found
using any combination of features, and all suffered from the same problem as the
four main features: good segmentations were too parameter-specific. Likewise,
solely grey-scale features did not yield any improvement. 

\begin{figure}[t]
\centering
\subfloat[]{\fbox{
\includegraphics[width=0.155\textwidth, clip,
				 trim={0 0 0 4}]{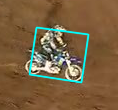}%
\label{fig:results_goodbad_vibe_1}}}
\hfil
\subfloat[]{\fbox{%
\includegraphics[width=0.155\textwidth, clip,
				 trim={20 20 20 20}]{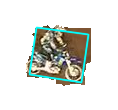}%
\label{fig:results_goodbad_vibe_2}}}
\hfil
\subfloat[]{\fbox{%
\includegraphics[width=0.155\textwidth, clip,
				 trim={20 20 20 20}]{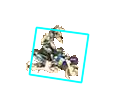}%
\label{fig:results_goodbad_vibe_3}}}
\caption{SBBM: (a) Original image. (b) Cross validated segmentation 
($\delta = 0.1$, $\nu = 1.0$). (c) Best possible segmentation 
($\delta = 0.3$, $\nu = 0.1$).}
\label{fig:results_goodbad_vibe}
\end{figure}

\paragraph{SBBM}
Similarly to the OC-SVM, SBBM also demonstrated the same problem of being too
parameter dependent. We performed the same comparison of the cross-validated 
segmentations to the best possible segmentations, an example of which can be 
seen in Fig.~\ref{fig:results_goodbad_vibe}. The technique was capable of 
creating good segmentations on some of the test images, e.g.\
Fig.~\ref{fig:results_goodbad_vibe_3}, but had a greater range of variation as
to how successful the best possible segmentations were, similar to its 
interquartile range in Fig.~\ref{fig:performance_box_whisker_all}.

\begin{figure}[t]
\vspace{-10pt}
\centering
\subfloat[]{\fbox{
\includegraphics[width=0.155\textwidth, clip,
				 trim={20 0 20 0}]{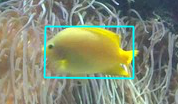}%
\label{fig:results_alphamatting_good_1}}}
\hfil
\subfloat[]{\fbox{%
\includegraphics[width=0.155\textwidth, clip,
				 trim={30 7 30 8}]{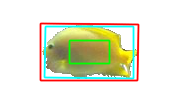}%
\label{fig:results_alphamatting_good_2}}}
\hfil
\subfloat[]{\fbox{%
\includegraphics[width=0.155\textwidth, clip,
				 trim={30 7 30 8}]{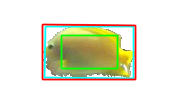}%
\label{fig:results_alphamatting_good_3}}}
\caption{Alpha matting: (a) Original image. (b) Cross validated 
($\rho^- = 0.8, \rho^+ = 1.2, \tau = 0.85, \lambda = 10^{-2}$) and (c) best
possible segmentations ($\rho^- = 0.6, \rho^+ = 1.1, \tau = 0.85, 
\lambda = 10^{-2}$), with labelled \textcolor{cyan}{original}, \textcolor{red}
{expanded}, and \textcolor{ForestGreen}{contracted} BBs.}
\label{fig:results_alphamatting_good}
\end{figure}

\begin{figure}[t]
\centering
\subfloat[]{\fbox{
\includegraphics[width=0.155\textwidth, clip,
				 trim={5 10.75 5 10}]{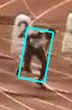}%
\label{fig:results_alphamatting_bad_1}}}
\hfil
\subfloat[]{\fbox{%
\includegraphics[width=0.155\textwidth, clip,
				 trim={10 20 10 15}]{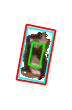}%
\label{fig:results_alphamatting_bad_2}}}
\hfil
\subfloat[]{\fbox{%
\includegraphics[width=0.155\textwidth, clip,
				 trim={10 20 10 15}]{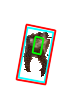}%
\label{fig:results_alphamatting_bad_3}}}
\caption{Alpha matting: (a) Original image. (b) Cross validated parameters
($\rho^- = 0.8$, $\rho^+ = 1.2$, $\tau = 0.84$, and $\lambda = 10^{-2}$)
and (c) changing parameters to $\rho^- = 0.9$ and $\tau = 0.6$, and 
translating the inner BB up by 9 pixels.}
\label{fig:results_alphamatting_bad}
\end{figure}

\paragraph{LBDM} The alpha matting technique achieved a much higher
segmentation accuracy for both $\phi_{all}$ and $\phi_{bb}$, as can be seen clearly in 
Fig.~\ref{fig:performance_box_whisker}, and improved on predicting the entire
BB as the object by a considerable margin. It tended to perform very
well in cases where the contracted BBs, green in 
Fig.~\ref{fig:results_alphamatting_good}, contained solely the object. However,
when this region contained background pixels, as is the case in 
Fig.~\ref{fig:results_alphamatting_bad_2}, these may be labelled as belonging to
the object and propagated outwards. As Fig.~\ref{fig:results_alphamatting_bad_3}
shows, shrinking and translating the inner BB to include only the
object improves the performance, but clearly this is not feasible in practice.

The group of objects whose centre of mass was positioned in some other location 
than the centre of the BB comprised almost completely of people. As 
people do not typically stand with their legs close together, and almost never 
do when in motion, we looked at the cross-validated performance of videos that 
contain humans (27) and those that do not (33). The average object segmentation 
performance $\phi_{all}$ for humans and non-humans was 0.718 and 0.800 
respectively. Comparing this to the average performance across all videos 
(0.763) shows that segmenting humans is a harder than average task. This is in 
contrast to segmenting non-human objects, appearing to be a simpler task, as
these typically are more compact. If one is able to detect the object to be
tracked is human-like, using a combination of shrinking the inner BB
further and translating it upwards, along with using a lower threshold $\tau$
for the amount of object likely in the BB, should improve performance
somewhat.

\begin{figure}[t]
\vspace{-12pt}
\centering
\subfloat[]{\includegraphics[width=0.235\textwidth, clip, trim=10 0 0 10]{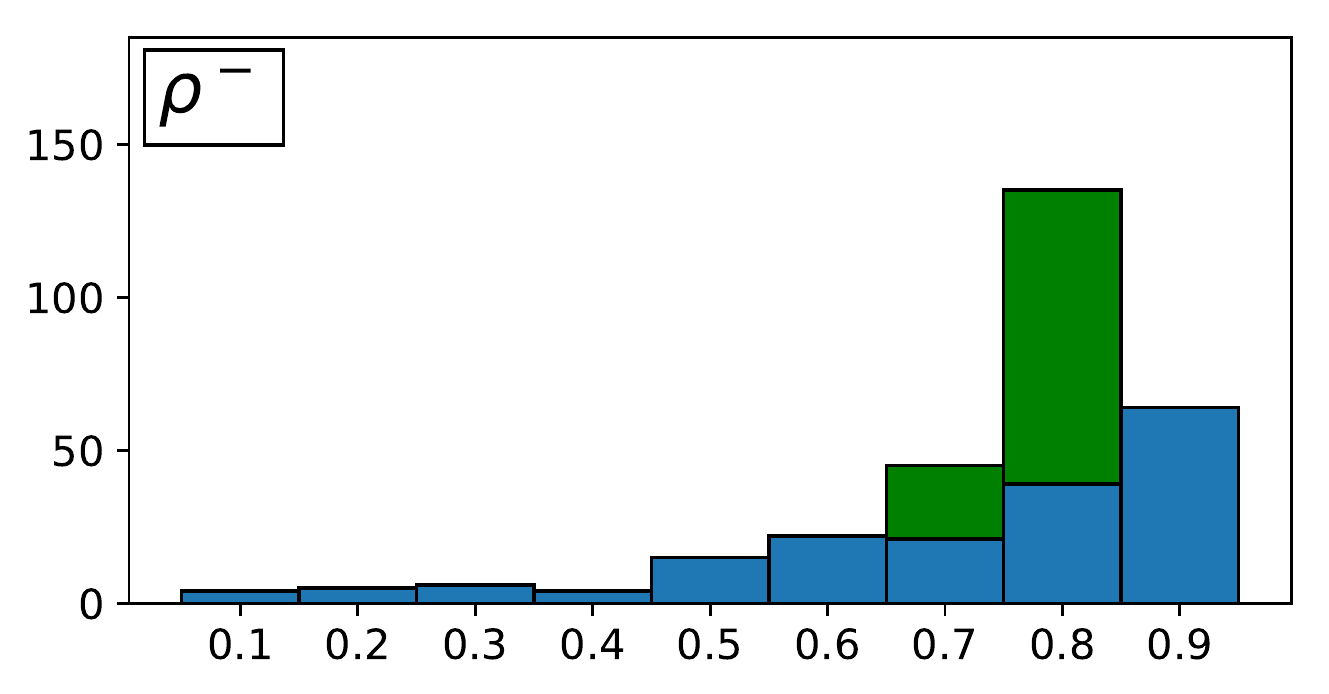}%
\label{fig:results_alpha_matting_param_dec}}
\hfil
\subfloat[]{\includegraphics[width=0.235\textwidth, clip, trim=10 0 0 10]{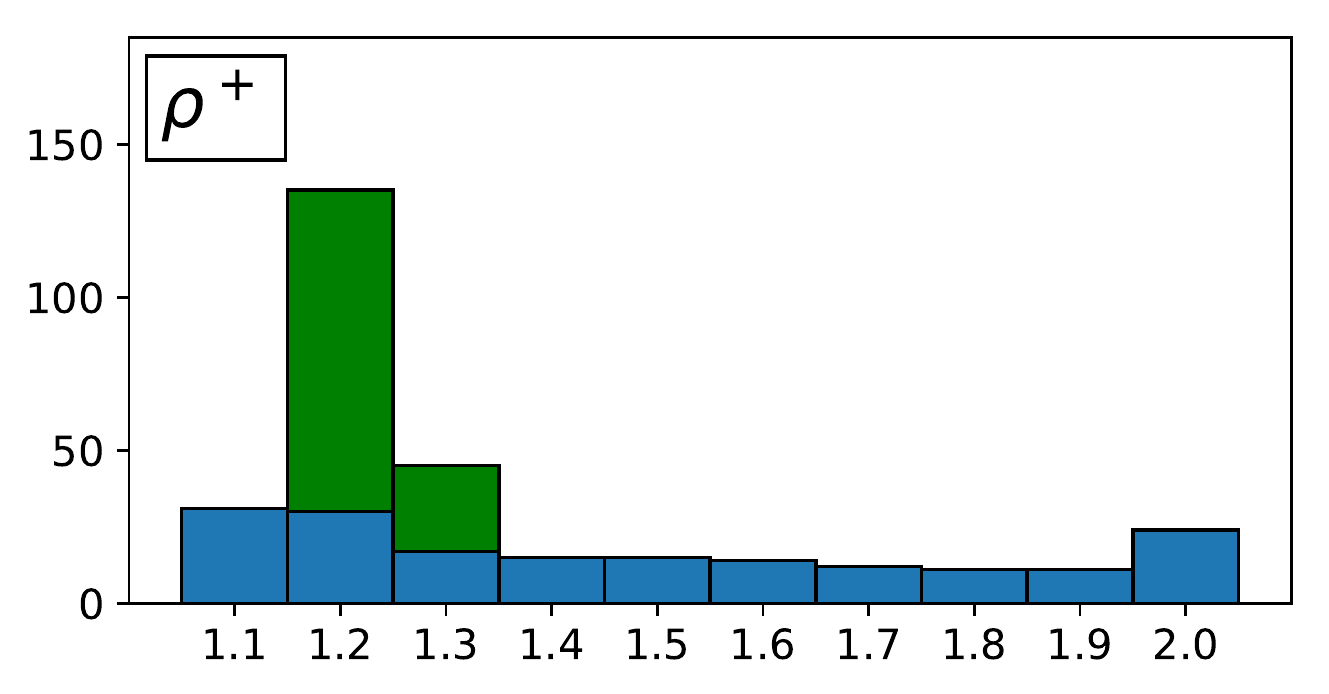}%
\label{fig:results_alpha_matting_param_inc}}%
\\[-2ex]
\subfloat[]{\includegraphics[width=0.235\textwidth, clip, trim=10 0 0 10]{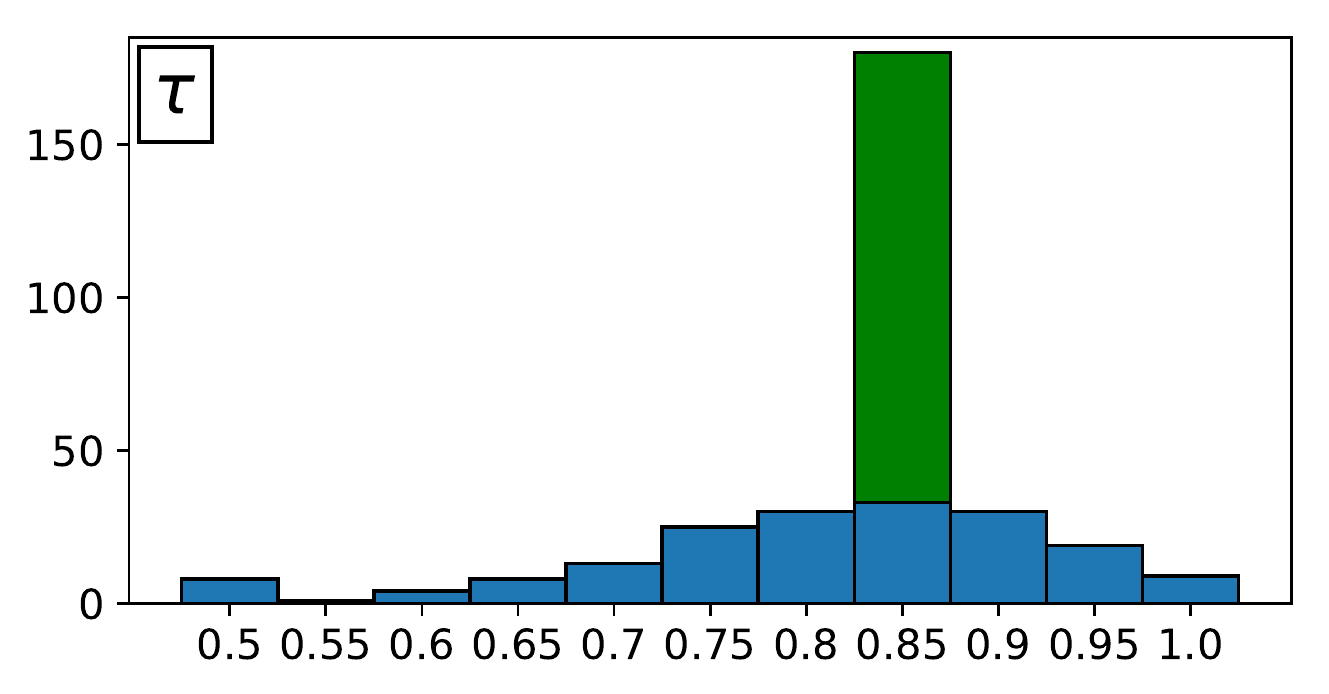}%
\label{fig:results_alpha_matting_param_bbfill}}
\hfil
\subfloat[]{\includegraphics[width=0.235\textwidth, clip, trim=10 0 0 10]{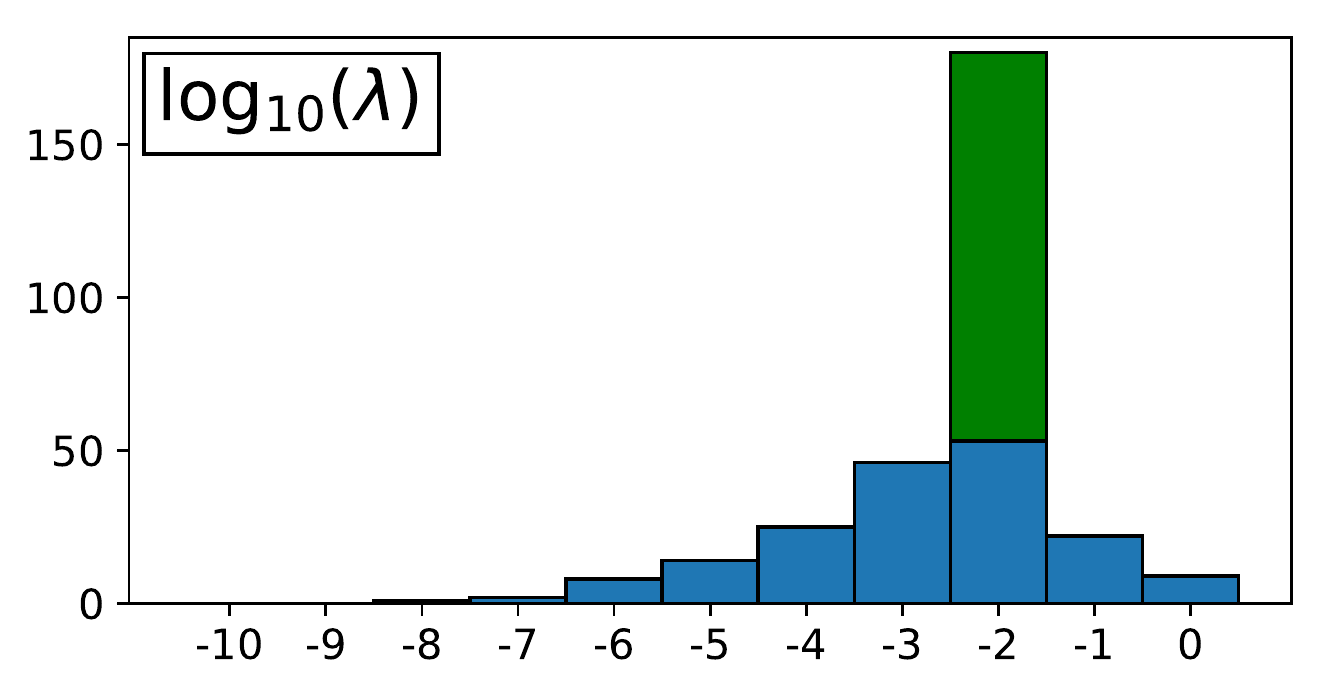}%
\label{fig:results_alpha_matting_param_lambda}}
\caption{Distribution of the \textcolor{blue}{optimal} and
\textcolor{ForestGreen}{cross-validated} alpha matting parameters for the 
$\phi_{all}$ measure. (a) $\rho^-$, (b) $\rho^+$, (c) $\tau$, and (d) 
$\log_{10}(\lambda)$.}
\label{fig:results_alpha_matting_param}
\end{figure}

Fig.~\ref{fig:results_alpha_matting_param} shows histograms of the parameters 
which gave the highest performance for each tested frame. The extent to which
the BB is shrunk to define scribble mask is controlled by $\rho^-$. 
As Fig.~\ref{fig:results_alpha_matting_param_dec} shows, for the majority of
images a large shrinkage (small scribble mask) is optimal. This helps to avoid
portions of the background, particularly for humans and other non-compact 
objects. Fig.~\ref{fig:results_alpha_matting_param_inc} shows a wide variation
in the optimal BB expansion parameter $\rho^+$, indicating that in
some circumstances it is helpful to include a large part of the 
adjoining regions, particularly for objects that extend outside the BB. 
However, the cross-validated optimum suggests an expansion of around 20\%.

The values (Fig.~\ref{fig:results_alpha_matting_param_bbfill}) of the fraction 
of the BB that should be filled by the segmentation, $\tau$, relate 
directly to the specific problem investigated. The benchmark \cite{VOT2016}
states that there should be no more than 30\% of pixels in the BB
containing background, which corresponds well to the region of most optimal
parameters, 0.75 to 0.90 in Fig.~\ref{fig:results_alpha_matting_param_bbfill}.
Fig.~\ref{fig:results_alpha_matting_param_lambda}, showing the optimal value of
the shrinkage term in the ridge regression, indicates a value between the two
values recommended by the technique's authors ($\lambda = 10^{-1}$ in 
\cite{LBDM} and $\lambda = 10^{-7}$ in their published code). This may indicate
that the parameter is more problem specific than they anticipated, and may be
related to the size of the features used or complexity of the image.

The alpha matting experiments were also repeated with the neighbourhood 
$\mathcal{N}_i$ of each pixel defined as a $7 \times 7$ window. This resulted in
approximately the same performance as using the $3 \times 3$ window, with 
$\phi_{all} = 0.760$ and $\phi_{bb} = 0.801$. \cite{LBDM} recommended the use
of a $7 \times 7$ neighbourhood size to give a more stable image matte, although
their images, and therefore the number of unlabelled pixels, were much larger
than here. The computational expense is dominated by solving 
\eqref{eqn:alpha_matting_thingtosolve} for $\bb{\alpha}^*$. With 
$7 \times 7$ windows ($m = 48$) it is approximately forty times more expensive
than for $3 \times 3$ windows ($m = 8$).

\section{Conclusion}
\label{sec:conclusion}

We have investigated three novel methods of determining the object to be tracked
within a limited BB. Two of these, OC-SVM and SBBM, depend on 
characterising the background image's colour/texture so that the foreground
object can be identified as different. Although both of these methods perform
well with tuned parameters, neither is robust enough to allow good
performance on novel videos. 

An alpha matting method (LBDM), which seeks to extend the foreground object from
a presumed \textit{scribble mask}, gives good performance over the VOT2016
dataset using parameters chosen by cross-validation. It is able to improve upon
the assumption of most trackers that the entire BB belongs to the
object, and should increase the initial performance of most tracking techniques.
The LBDM method relies on the assumption that the centre of the BB is
the object to be tracked and its performance is degraded for non-compact objects
for which this is not the case.

All of these methods suffer from the paucity of data and we expect improved
performance as higher resolution video becomes available. We note that the 
initialisation problem is significant not only at the start of tracking, but
also when re-initialisation is required during tracking.

Source code for all three methods is publicly available at: 
\url{http://github.com/georgedeath/initialisation-problem}.

\bibliographystyle{IEEEtran}
\bibliography{IEEEabrv,refs}
\end{document}